\documentclass[runningheads]{llncs}
\usepackage{graphicx}
\usepackage{booktabs}
\usepackage{subfigure}
\usepackage{float}

\begin{document}
\title{How language of interaction affects the user perception of a robot\thanks{This research was funded in part by the National Science Centre, Poland, under the OPUS call in the Weave programme under the project number K/NCN/000142, by the Czech Science Foundation grant no. 22-04080L, and by the CTU Student Grant Agency (reg. no. SGS21/184/OHK3/3T/37)}}
\titlerunning{The Impact of Language on Robot Perception}
%

\author{{Barbara Sienkiewicz}\inst{1}\orcidID{0009-0008-1977-5806} \and {Gabriela Sejnova} \inst{2}\orcidID{0000-0002-0045-6425}\and {Paul Gajewski} \inst{3} \orcidID{0000-0003-0931-2476}\and {Michal Vavrecka} \inst{2}\orcidID{0000-0002-0152-2682} \and
Bipin Indurkhya\inst{1}\orcidID{0000-0002-3798-9209}}

\authorrunning{B. Sienkiewicz et al.}

\institute{Cognitive Science Department, Jagiellonian University, Krakow, Poland \and Czech Institute of Informatics, Robotics and Cybernetics, Czech Technical University, Prague, Czechia \and Institute of Computer Science, AGH University of Science and Technology, Krakow, Poland}

\maketitle 

\begin{abstract}
Spoken language is the most natural way for a human to communicate with a robot. It may seem intuitive that a robot should communicate with users in their native language. However, it is not clear if a user's perception of a robot is affected by the language of interaction.

We investigated this question by conducting a study with twenty-three native Czech participants who were also fluent in  English. The participants were tasked with instructing the Pepper robot on where to place objects on a shelf. The robot was controlled remotely using the Wizard-of-Oz technique. We collected data through questionnaires, video recordings, and a post-experiment feedback session. The results of our experiment show that people perceive an English-speaking robot as more intelligent than a Czech-speaking robot (z = 18.00, p-value = 0.02). This finding highlights the influence of language on human-robot interaction. Furthermore, we discuss the feedback obtained from the participants via the post-experiment sessions and its implications for HRI design.
\end{abstract}


\section{Introduction}\label{sec1}
Social robots that interact with ordinary people are becoming increasingly commonplace. It is important that this interaction be facilitated naturally, similar to human-human communication \cite{thomaz2008teachable}.
Spoken language is one such natural medium of communication\cite{marge2022spoken}. Speech-based interfaces allow a social robot to be used effectively and flexibly in many situations\cite{bainbridge2011benefits}, such as tutoring children\cite{marge2022spoken}.

Towards this goal, we need to study how language and speech affect people's perception of a robot. This applies to all voice-based human-robot interfaces, whether they are on land, sea, air or in cyberspace.
Some aspects of speech have already been investigated: for example, vocal prosody\cite{breazeal2001emotive}, the melody of speech\cite{fischer2019speech}, or formal/informal speech\cite{steinhaeusser2022designing}.
However, the effect of language itself (native language as opposed to a non-native language like English) has not yet been studied. We hypothesize that there is a difference in the perception of the robot when the interaction is in the native language or a familiar foreign language. We conducted an empirical study to study this effect.This article is an expanded version of a brief study that is published elsewhere \cite{barb}. In this article, we provide more details of the related research, experimental set up, and a more through data analysis and discussion.

\section{State of the art}\label{sec2}

\subsection{Human perception of a robot}
A user's perception of a robot depends on several factors, e.g. appearance \cite{song2017expressing}, behavior\cite{tan2020relationship}, and their correlation \cite{goetz2003matching}. However, these factors may vary with respect to the application area\cite{sheridan2016human}. 

A user's perception of robots is influenced by the context of interaction: e.g., the perception differs when the robot is a teacher \cite{polishuk2018elementary} as opposed to when the user is teaching the robot \cite{thomaz2006reinforcement}. Moreover, people with a prior knowledge about robots tend to perceive them differently than those without \cite{hall2014perception}. Finally, cultural and social context also plays a role in the perception of robots: e.g., people in Japan have more positive attitudes towards robots than in Europe\cite{sone2016japanese}. 

\subsection{Speech, language, and Ethnicity in HRI}
Rapid advances in computer-based speech understanding (e.g., Apple’s SIRI) suggest that it will become easier to command a robot using speech \cite{sheridan2016human}.
Marge {\it et al.}\cite{marge2022spoken} propose directions for further development and improvement of spoken language interaction between humans and robots. Their recommendations cover topics such as multimodal communication, dialogue management, and user-centred design, aiming to create intuitive and effective communication systems.
T. Takahashi et al.\cite{takahashi2021melodic} argue that incorporating emotional expressions into a robot's speech makes it more human-like and easier to talk to.
Cultural aspects play a dominant role in speech-based human-robot interaction.
A study conducted in Qatar with native English and Arabic\cite{salem2014marhaba} speakers found that the Arabic-speaking participants held a more positive perception of the robot and anthropomorphized it more than the English-speaking participants. Another study \cite{trovato2013cross} found that the Japanese prefer a Japanese robot and feel discomfort while interacting with an Egyptian robot, and the opposite is true for the Egyptians. Such results show the importance of robot ethnicity in improving human acceptance during human-robot social interactions.

Other studies show  differences in interaction with robots when using different accents \cite{andrist2015effects},\cite{torre2021trust}.
Language is one factor that contributes to ethnic differences; other factors include race, skin color, historical background, and religion\cite{nam2020trust}. A robot’s perceived ethnicity has a significant effect on human trust and robot's perception\cite{gong2008boundary}. Manipulating the robot's language allows us to change its ethnicity and how it is perceived by the user. 

\subsection{Role of language in human-human communication}
Here we briefly comment on the effect of using the mother tongue or a foreign language on human-human communication. Language is a primary means of human communication, allowing individuals to express their thoughts, feelings, and ideas to others. When communicating with others, individuals typically use their mother tongue, which is the language they learned first as a child. Using one's mother tongue influences interaction, as it is often associated with a sense of familiarity, comfort, and cultural identity\cite{snow1990rationales}. However, communicating in a foreign language can also have both positive and negative effects on the interaction. On one hand, using a foreign language can provide opportunities for cross-cultural communication and learning. On the other hand, it can create communication barriers and misunderstandings due to language proficiency and cultural differences\cite{ting2018communicating}. In multicultural societies, English serves as a lingua franca, facilitating communication across diverse backgrounds\cite{henderson2011does}. However, due to its limited shared understanding, it can sometimes create confusion and uncertainty. Additionally, interacting in English lingua franca involves some vulnerability and potential risk that are not present when communicating within a shared linguistic community that shares social meaning\cite{henderson2011does}. Recent research has also shown that the phenomenon of false memory in bilingual people is affected by whether they are required to match the language in which the initial information was given and the language in which they were asked to recall the information\cite{Beato2023}.

Given all these factors related to how the language of interaction affects human-human communication, we would like to explore how the language of communication affects human-robot interaction.

\section{Experiment}\label{sec3}
\subsection{Objective}
The research question we focused on was how a human's perception of a robot is influenced by the language of interaction. We compared the interaction in the native language of the speaker (Czech) with the interaction in a well-known foreign language (English). 
Participants were asked to teach a robot how to organize objects on a desk and shelves above it. Each participant interacted with the robot under two conditions (the order of the conditions was randomized):
\begin{itemize}
\item Condition A: The participant and the robot spoke to each other in English.
\item Condition B: The participant and the robot spoke to each other in Czech.
\end{itemize}
We hypothesized that the participants' perception of the robot would differ depending on the language of interaction, and they would prefer interacting in their native language. The collected data combined our observations of the participants' behaviour during the interaction, the participants' self-reported answers to a questionnaire after each interaction, and post-test interviews with each participant. 

\subsection{Task}
There is growing interest in the integration of robots into domestic environments \cite{kim2019control}, which poses a number of challenges. As each home and its inhabitants are unique, robots must be able to learn how to operate in different environments and non-expert users should be able to explain the robot its task. 

\begin{figure}[h!]
\centerline{\includegraphics[width = 0.4\textwidth]{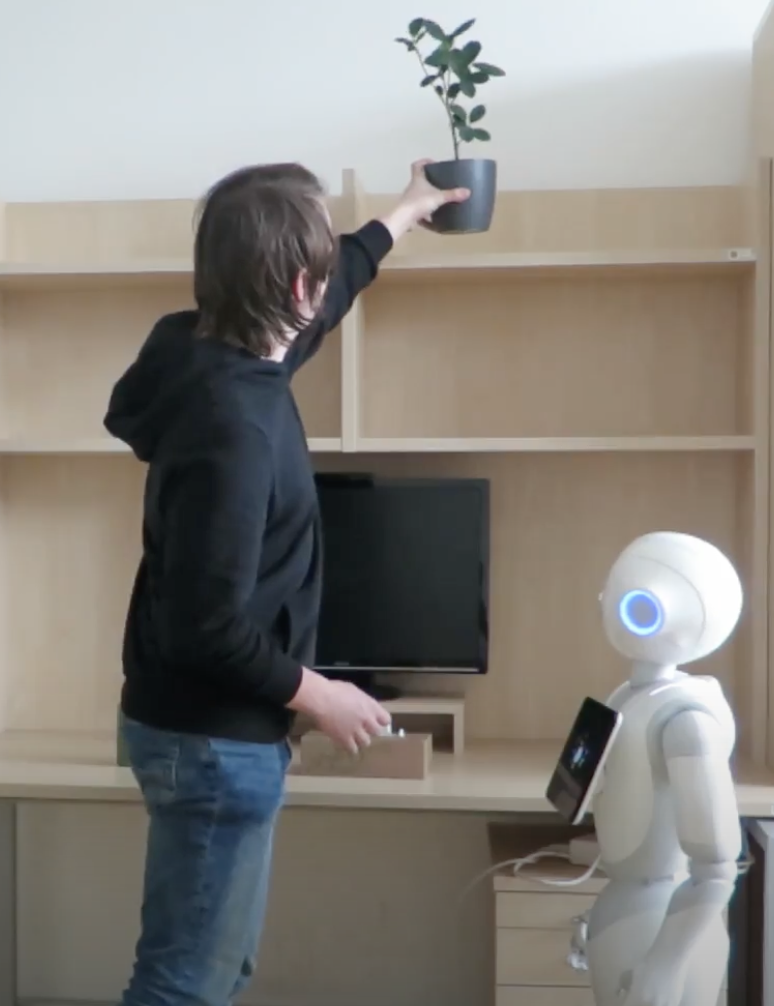}}
\caption{The user demonstrating to the robot where to place an object}
\label{Fig.Interaction}
\end{figure}

\begin{figure}[htbp]
    \centering
    
    \subfigure[Objects used in condition A (interaction in English)]{\includegraphics[width=0.45\linewidth]{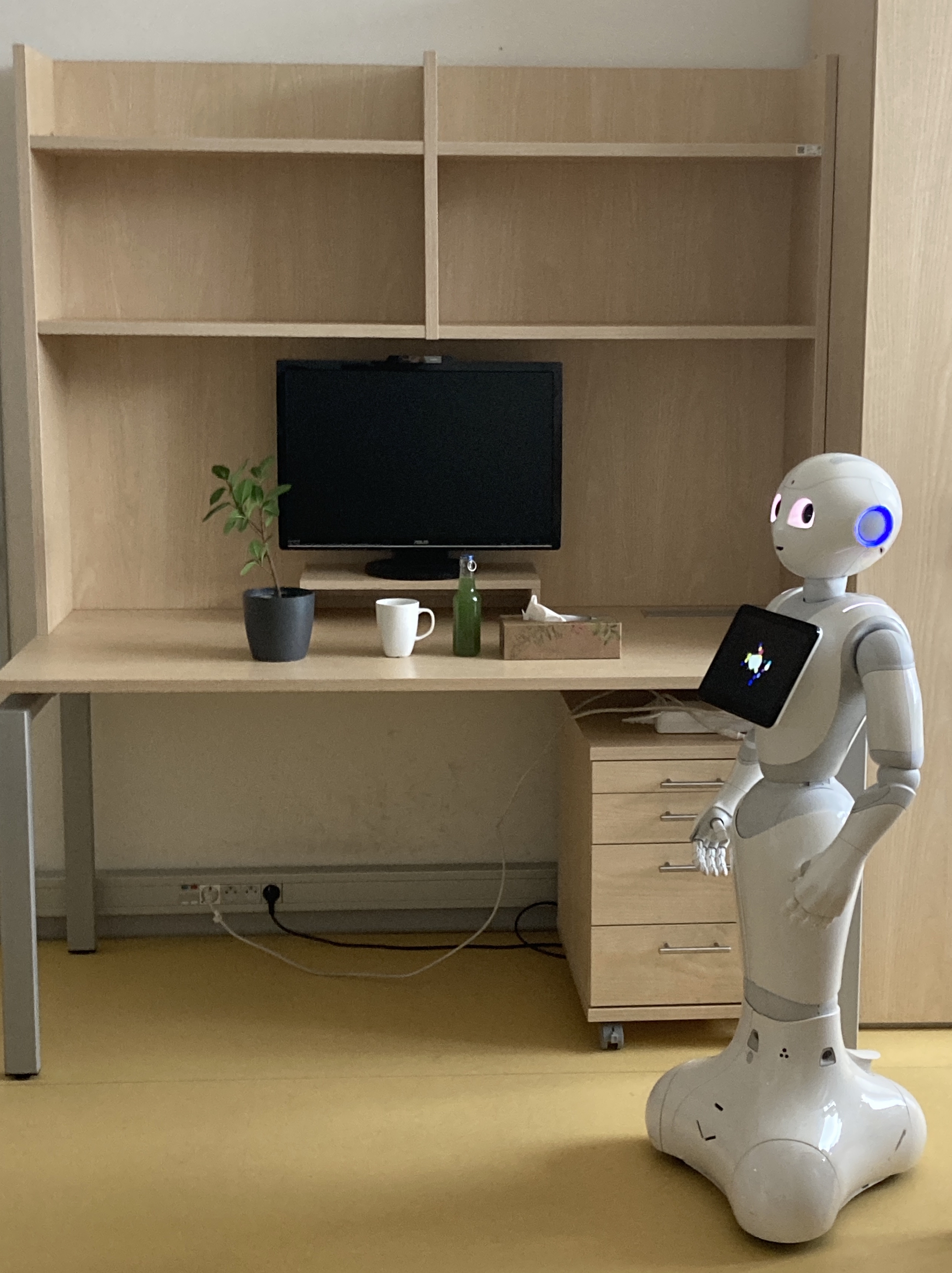}}
    \hfill
    \subfigure[Objects used in condition B (interaction in Czech)]{\includegraphics[width=0.45\linewidth]{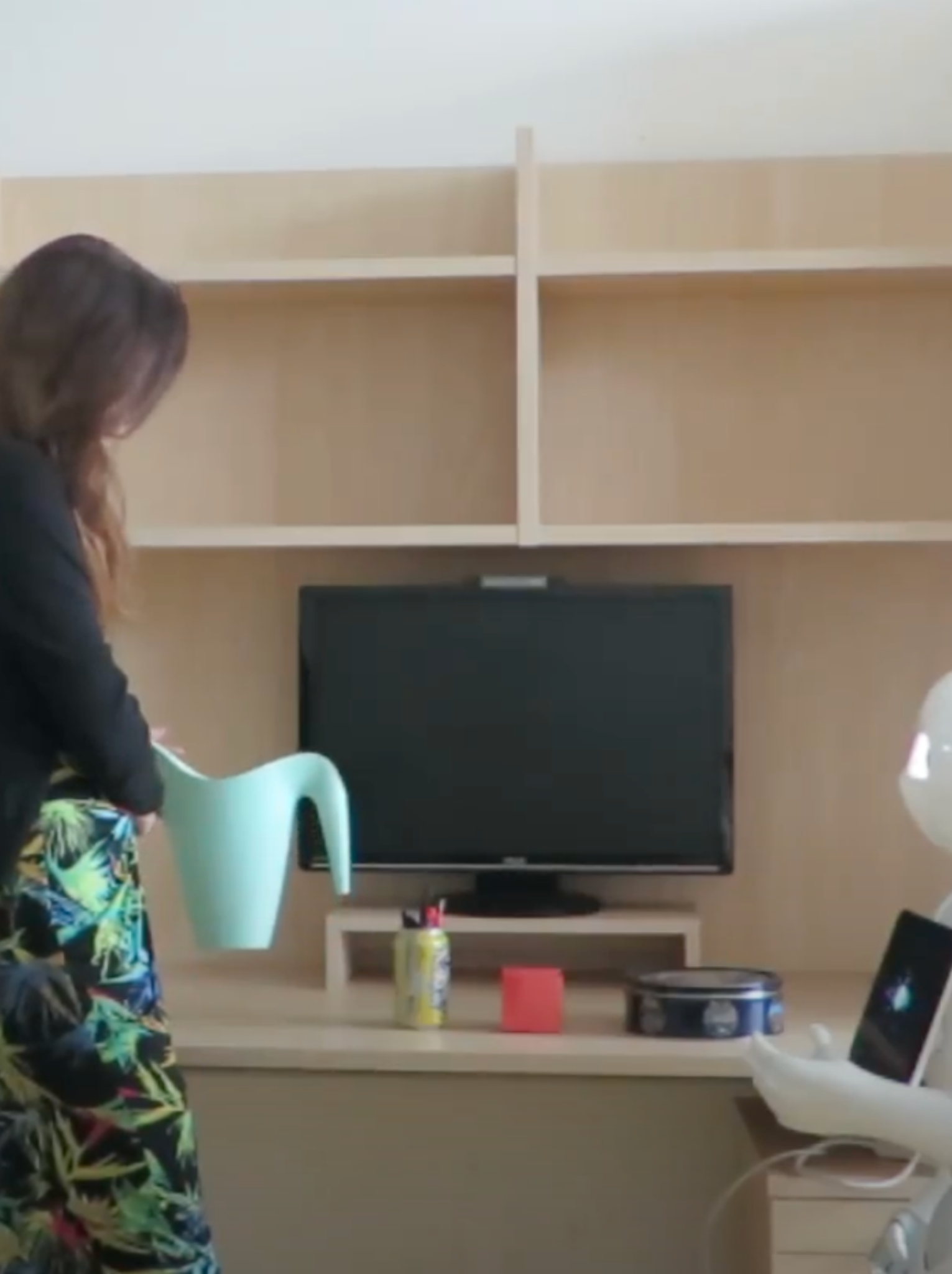}}
    \caption{Set up of the experiment}
    \label{fig2:side_by_side}
\end{figure}

To address these requirements, we designed a real-life scenario in which a user has to teach a robot how to organize objects to tidy up. By instructing the robot to clean and organize their desk, the participant would potentially be relieved of this responsibility in the future, which is a great motivation for accomplishing the task. Moreover, the task did not require specialized domain knowledge or technical jargon, making it easily comprehensible and feasible to demonstrate. Participants were instructed to treat the desk as their own and teach the robot the desired organization of the objects, considering their personal preferences (Fig.1). The concept of {\it teaching} was defined as physically handling each object and placing it in the desired location while providing verbal descriptions of the actions, as it can be seen in Fig.\ref{Fig.Interaction}. Furthermore, participants were informed that the robot would remain stationary throughout the interaction, observing but not performing any actions. Fig.\ref{fig2:side_by_side} (a) shows the setup before the interaction started for condition A. 

To ensure the validity and reliability of the experiment, we employed two distinct sets of objects, one for each condition. In Condition A, the objects included a plant, a mug, a glass bottle and a tissue box (Fig.\ref{fig2:side_by_side} (b). For Condition B, the objects consisted of a metal cookie tin, a watering can, a cup with pens, and a red paper cube.

There were several rationales behind using different sets of objects. Firstly, it prevented any references or comparisons to previous conditions, ensuring that participants approached each teaching session independently. This minimized potential biases or influences that could arise from participants' previous experiences.

Secondly, the use of diverse objects enhanced the realism of the scenario, making it more representative of real-life situations. In real-world contexts, individuals encounter various objects with different shapes, sizes, and functions. By replicating this diversity in the participants' experimental setup, we aimed to capture the complexity of teaching a robot to organize objects in a practical manner. Another reason for using distinct objects in each condition was to facilitate language consistency. As the {\it wizard} (the person operating the robot) was physically present in the same room as the researcher and the participant, they could not ask clarifying questions or show emotions on their face. By associating specific objects with a particular language, it was easier for the wizard to recall and maintain the appropriate language throughout the interaction. This ensured clarity and minimized potential language-related errors or confusion during the teaching process.

\subsection{Experiment set up}
The experiment took place in a room at the Czech Institute of Informatics, Robotics, and Cybernetics, where researchers frequently work. 

\begin{figure} [H]
\centerline{\includegraphics[width = 0.4\textwidth] {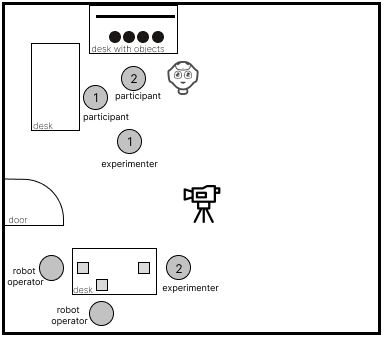}}
\caption{Bird's eye view of the study setup }
\label{Fig.Schematic of the study setup }

\end{figure} Figure \ref{Fig. Schematic of the study setup } provides a
bird's eye view of the experimental setup. During the initial stage and questionnaire sessions, the participant was seated at Location 1, and was asked to stand at Location 2 for the interaction phase. The experimenter stood at Location 1 in the initial phase and at Location 2 in the interaction phase to give the participant freedom and privacy with the robot.
There were two robot operators involved in the experiment: one was responsible for controlling the robot's speech, while the other operated its movement. Additionally, in the other part of the room (shown as white space on the right side of the figure), two other researchers were working on their computers. The participant did not expect the robot operators to be directly involved in the experiment, as several people were present in the room. Besides the participant and the experimenter, everyone else was quiet. The camera covered the desk at which the participant was sitting, the desk with the objects, and the robot. 

\subsection{Robot Operation}
First the humanoid robot used in the experiment is described with its hardware and software specifications. Then, we describe our program designed to operate the robot using the Wizard-of-Oz method.

\subsubsection{Humanoid robot}
We used Pepper, a humanoid robot developed by SoftBank Robotics and released in June 2014 (see Fig.\ref{fig2:side_by_side} (a)). Pepper is equipped with two identical video cameras on the forehead and a 3D video camera. It is capable of facial recognition and analysis of human expressions. In addition, four microphones in the robot's head allow it to analyze speech and vocal tones, and to detect the direction of incoming sounds.

\begin{figure*}[h!]
\centerline{\includegraphics[width = 0.8\textwidth] {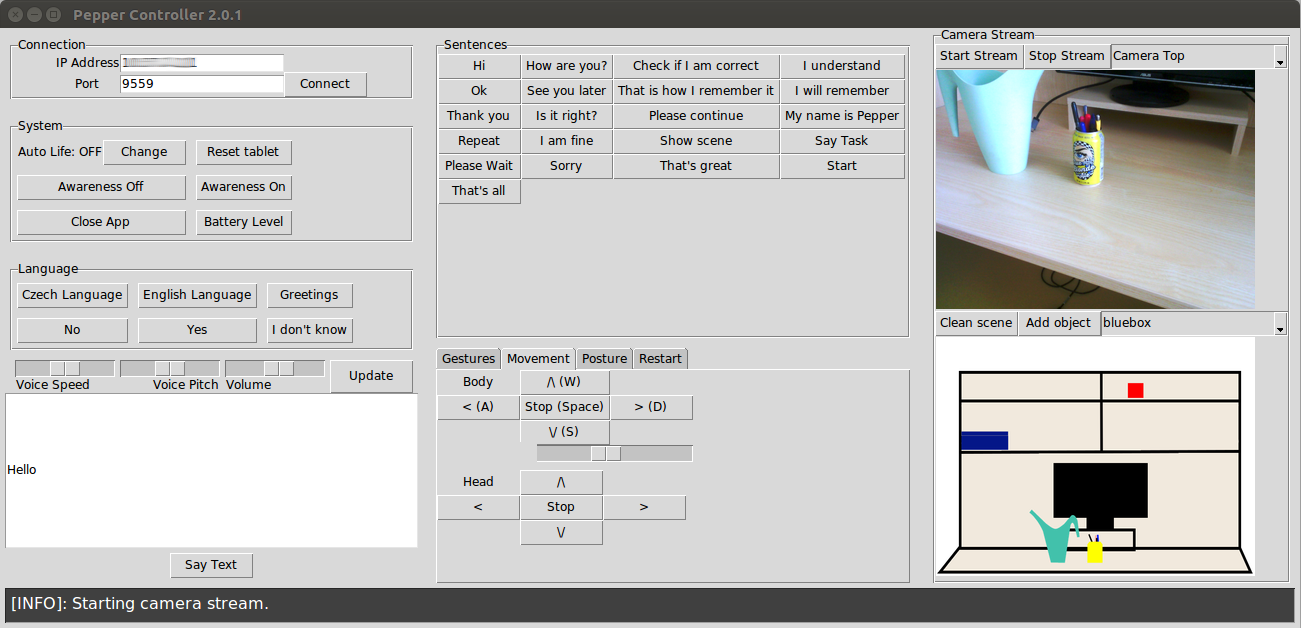}}
\caption{GUI for operating the robot using the Wizard-of-Oz method. We control the language (left-middle), speaking (top-middle), and head movements (bottom-middle), and display the current scene on the tablet when needed (bottom-right). We can also watch the scene using the robot's head camera (top-right).}
\label{peppergui}
\end{figure*}

Pepper was controlled by the NAOqi operating system, which is a programming framework for developing applications on the robot. NAOqi addresses common robotics requirements such as parallelism, resource management, synchronization, and event handling. NAOqi functions can be called in C++ or Python, and a graphical environment called Choregraphe is available to create complex behaviors for Pepper. Our research team developed a Python API wrapper around the NAOqi framework, which was used in this study.

The Pepper class, the main component of our Python API, establishes a connection to the robot instance using a specified IP address and port. Once the connection is made, it provides tools to interact with the robot, such as moving its arms, legs, and head, changing its LED lights, playing sounds and speech, and accessing its sensors and cameras. The API also allows the creation of custom behaviors and applications using Python code that can be run on the robot.

\subsubsection{Wizard of Oz}
The Wizard-of-Oz method\cite{riek2012wizard} was implemented using a GUI built on top of our Python API described above. After establishing a connection with the robot, we can control it by clicking the corresponding buttons in the GUI. Each button is mapped to a Python function in the Pepper class. The GUI overview is shown in Fig. \ref{peppergui}. Its main components are as follows:

\begin{itemize}
    \item Connection panel - to connect with the robot via its IP address
    \item System - to control the level of autonomous movement of the robot and reset any current process
    \item Language - enables to switch between Czech and English language and to control voice volume, speed and pitch
    \item Sentences - each button is mapped to a sentence that the robot says in the selected language
    \item Movement - enables head movement
    \item Camera Stream - displays what the robot sees using its top head camera
    \item Scene scheme - the operator can edit the current scene by adding objects as they are arranged by the participant. The "Show Scene" button then displays this image on Pepper's tablet
\end{itemize}

Though the robot can move around on its wheels, for this experiment, it remained in one place and only moved its head in a natural way to give the impression of observing the actions of the participant.
Voice parameters (such as volume, speed, and pitch) remained fixed throughout the experiments. Besides its head, the robot moved its arms and hands to gesticulate while speaking. After each object shown by the participant, the robot confirmed that he understood ("Okay", "I see", "I understand") with appropriate gestures and head movements.

\subsection{Questionnaires}
We used the following questionnaires and tests:
\begin{itemize}
    \item Pre-test: To collect demographic data and information about the participants' experience with programming and familiarity with robots. 
    \item English test: To verify the level of English, each participant was asked to complete a short English test for B2/C1. 
    \item Godspeed Questionnaire(\cite{bartneck2009measurement}, \cite{bartneck2023godspeed}): Standardised questionnaire to measure the perception of robot in 5 subscales: anthropomorphism, animacy, likeability, perceived intelligence, perceived safety. 
    \item Post-interaction questionnaire: After each condition we asked the participant how easy it was to check the robot's knowledge, whether the robot made any mistakes, and whether the robot was capable of retaining the knowledge. This questionnaire was designed to explore the participant's reasoning. It also allowed us to compare the actual errors with the ones noticed by the participants. Each participant filled two questionnaires, one after each condition.
    \item Post-test questionnaire: This was designed to get the participant's overall feedback to evaluate the study methodology. 
\end{itemize}

All questionnaires were in English and Czech, in order not to bias any language.
\subsection{Participants}
Twenty-three participants (9 women, 14 men), aged (14-72 yrs,  M = 32.04, SD = 14.22), were recruited through emails and web announcements. Informed consent was obtained from all the participants above 18 yrs. and from their parents for those under 18 yrs. (four participants). 
To avoid language skills as a confounding factor, the requirement to participate in the study was proficiency in both English and Czech. 

\subsection{Procedure}\label{AA}
Each participant interacted with the robot individually. The experiment took about 20-30 minutes. The procedure was as follows:

\begin{enumerate}
\item 
 The participant was seated at a desk with their back to the robot. At first, the participant took the English test; they were told that their test performance would not impact their interaction with the robot. Then the participant read and signed the consent form, followed by filling the pre-test questionnaire. The English test was given first, and was followed by other tasks to maximize the gap between the test and the interaction for avoiding bias.
 
\item 
The participant was asked to move near the robot and the task was explained. If (s)he had any questions or hesitations, explanations were provided without disclosing the purpose of the study or how the robot works. The participant was asked to use a particular language (Czech or English) before each interaction. The order of conditions was random.

\item
The camera was turned on, and the robot initiated the interaction, which lasted about 2-3 minutes. Then, the camera was turned off, and the participant returned to the desk to complete the Godspeed and post-interaction questionnaire, while the objects in the scene were changed.

\item
The second round of interaction and questionnaires followed similarly to the first round, except that the language of interaction was switched and the objects to be arranged were different. 

\item
The participant was asked to provide feedback on the paper and was compensated 200 Czk (about 10 euros) for their participation. The participant was then debriefed about the purpose of the study and the Wizard-of-Oz methodology, revealing the robot operators. As this part was not recorded, participants' comments from this part were written down in a file after the participant was gone.
\end{enumerate}

\section{Results and Discussion}\label{sec4}
Data from one participant was excluded from the analysis due to numerous deficiencies in the questionnaires. One participant wrote in the survey his suspicions that the robot is controlled by a human, but the rest of 21 participants did not express any such doubts.
\subsection{Pre-test}
Most participants had no experience with humanoid robots, but some had knowledge of programming and robotics. The most common answers to questions about the robot's appearance were cute, friendly, child/childlike.
\subsection{English test}
We conducted an English test to verify fluency in English. All participants received 8 or more out of a possible 13 points in an English-level test between B2 and C1.

\subsection{Godspeed Questionnaire}
\begin{figure}[htbp]
    \centering
    
    \subfigure[Condition A: the English language]{\includegraphics[width=0.45\linewidth]{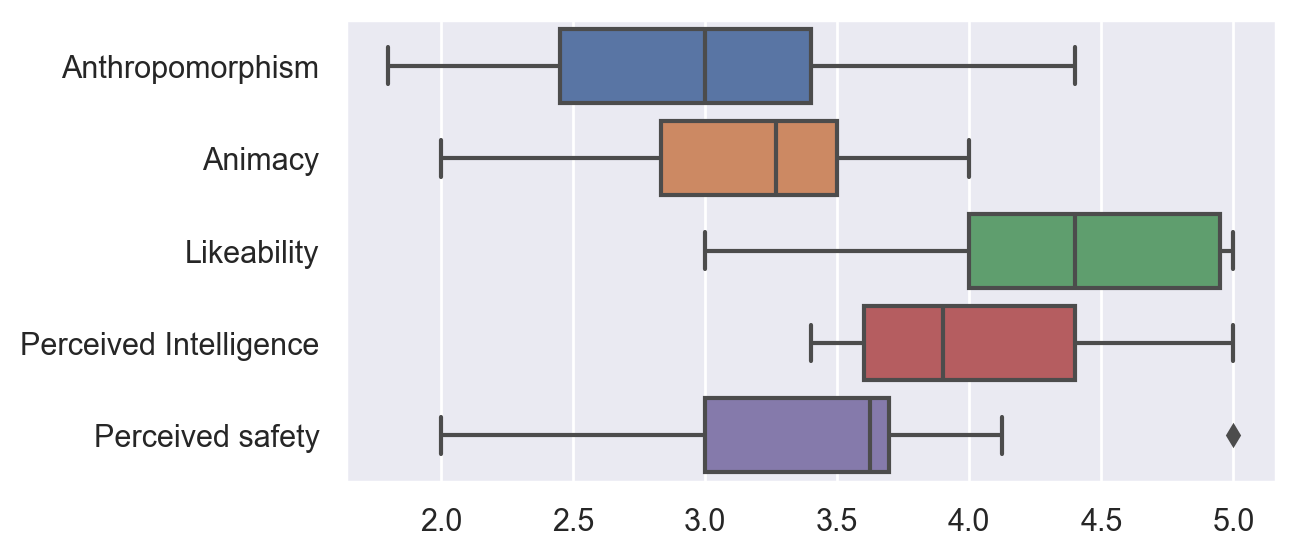}}
    \hfill
    \subfigure[Condition B: the Czech language]{\includegraphics[width=0.45\linewidth]{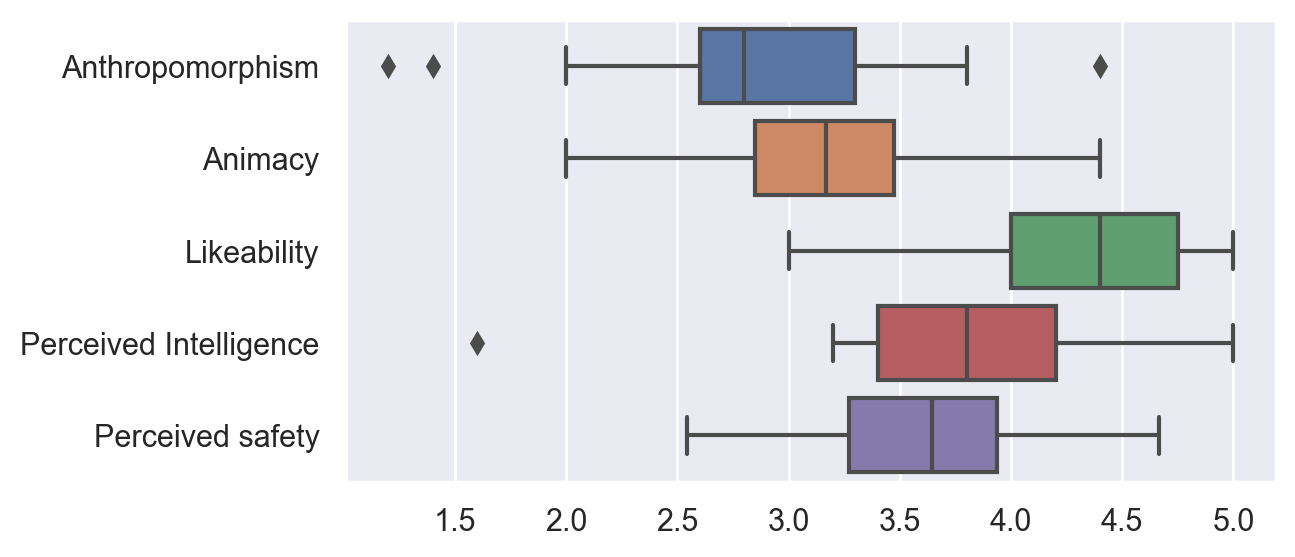}}
    
    \caption{Distribution of results in both conditions}
    \label{fig5:side_by_side}
\end{figure}

The Godspeed Questionnaire was used to measure participants' ratings of the robot in five sub-scales. Table\ref{tab:godpseed} shows the average scores, and the distribution of results is shown in Fig.\ref{fig5:side_by_side}.
We are aware of the criticism regarding the use of intelligence as a dependent variable \cite{weiss2015meta}, but we still considered it as a relevant outcome measure for our study. This criticism refers to the evaluation of operator intelligence during the Wizard-of-Oz methodology, rather than that of the robot. In our study, the same operator employed a set of predetermined sentences (in translation), consistently supervising the robot's speech and movements. 

Based on a previous study \cite{gombolay2016appraisal}, we used a nonparametric test to evaluate the results. For our within-subjects study we used the Wilcoxon test.
Only the subscale of perceived intelligence showed a significant result, z = 18.00, p-value = 0.02 ; an English-speaking robot is perceived as more intelligent than a Czech-speaking robot. 
A plausible explanation for the observed phenomenon is that when the robot communicates in a foreign language (English), participants may perceive it as more intelligent due to the increased difficulty in detecting errors compared to their native language. This language barrier could lead participants to allocate more cognitive capacity toward explaining the organization of objects, resulting in reduced attention toward the robot itself. Support for this hypothesis comes from the feedback questionnaire, where participants expressed their observations. For instance, one participant stated, "[...] when he speaks Czech, he was using words that I would never use in my daily life." Another participant remarked, "It seemed more lively when using English." Yet another participant commented, "For me, it was better in English as I could notice a slightly unusual wording or tempo of the sentences [in Czech]." These responses provide evidence supporting the notion that language choice influences participants' perceptions and attentiveness during human-robot interaction

\begin{table}[t]
  \caption{Average Scores in Godspeed Questionnaire}
  \label{tab:godpseed}
  \centering
  \begin{tabular}{lcc}
    \toprule
    \textbf{Subscale} & \textbf{\textit{English}} & \textbf{\textit{Czech}} \\
    \midrule
    Anthropomorphisation & 2.96 & 2.84\\
    Animacy & 3.16 & 3.15 \\
    Likeability & 4.3 & 4.32\\ 
    Perceived Intelligence & 4.04 & 3.79 \\ 
    Perceived safety & 3.44 & 3.60\\
    \bottomrule
  \end{tabular}
\end{table}

We investigated whether the sequential order of interactions influenced the ratings provided by the participants by comparing the results from the first interaction with those from the second, irrespective of the specific experimental conditions. Our analysis revealed no statistically significant differences between the two interactions. This suggests that the order of interactions did not have a significant impact on the outcomes observed in the study.

This result suggests a potential advantage in utilizing English language during robot interactions, as it appears to yield more favorable outcomes, which is desirable in the design of social robots for widespread adoption. However, it is important to acknowledge that this approach could restrict the participant pool to individuals proficient in English, as speaking a foreign language often generates feelings of anxiety\cite{horwitz2001language}. Hence, it is necessary to study this issue further: for example by conducting a study with native English speakers, and also with a larger sample. 

\subsection{Questionnaire after the interaction}
Based on the responses of the participants on a six-point Likert scale, the majority of the trials (37 of 44) were rated as easy (5) or very easy(6). Lower scores were observed in cases where the robot experienced technical issues, such as delays in response time or errors in visualization (6 out of 44 trials). Some participants considered the wrong orientation of the tissue box as an error, which was not expected. (The system was unable to display objects in different orientations.) This suggests that the system could be improved e.g. to rotate objects in future studies. 

\subsection{Insights from the feedback questionnaire}
The feedback questionnaire proved valuable in comprehending the participants' perceptions of the research. Notably, a majority of responses focused on exploring the robot's capabilities and evaluating its usability in human-robot interaction, with only one participant highlighting language as the primary concern. This suggests that the study effectively diverted attention from the language aspect, potentially influencing participants' expectations and perception of the research. It is important to note that bilingual individuals often encounter difficulty in recalling the language in which information was presented, without affecting their ability to recall the content. This phenomenon may have positively influenced the participant's response.

We included questions related to the experiment itself: clarity of instructions, areas of confusion, and potential improvements. Respondents expressed satisfaction with the study, frequently stating phrases such as "It was fun," "Everything was clear," and "I would like to have a robot like this at home." Two participants suggested that incorporating physical demonstrations of teaching the robot, rather than solely relying on verbal explanations, would have been beneficial. The deliberate omission of such demonstrations was intentional, as we aimed to observe participant attitudes without imposing specific instructional methods.

A participant commented that there were too many questionnaires, indicating the importance of appropriately adjusting the quantity to maintain participant engagement. When participants become excessively bored, they may not attentively read the questionnaires, resulting in uninformative data. To address this issue, we recommend including questionnaires in pilot studies to adjust their length and content. We also analysed the emotional aspects of the participants' responses.
One participant expressed feeling nervous due to the presence of many people in the room. This suggests making the experiment venue less crowded for future studies. Another participant expressed fear toward the robot's hands, highlighting the significance of individuals' attention towards robot appendages \cite{piazza2019century}. Although sometimes subjective and specific to individual participants, it is crucial to recognize that participants are humans with emotions and needs, not merely sources of data. We must remember that the primary objective of HRI is to understand human-robot interaction, which includes comprehending the human user's emotional responses \cite{marge2022spoken}.

\section{Conclusions}\label{sec5}
The results of the Godspeed questionnaire show that the participants perceived the English-speaking robot to be more intelligent than the Czech-speaking one. However, this result may be attributed to the fact that errors and unnatural behaviors are more difficult to be detected in a second language, as indicated by some participants in the feedback questionnaire. 
The limitations of the present study were: a small sample size, a diverse age range of the participants, and the absence of a comparison group comprising native English speakers fluent in Czech. We plan to address these issues in future research.

The importance of feedback questionnaires in HRI research has been highlighted by our research: they provide valuable insights into the participants' understanding and emotional responses towards the interaction and the study as a whole. 

\bibliographystyle{splncs04}
\bibliography{main}
\end{document}